\documentclass{article}
\usepackage{acra}

\usepackage{tabularx}
\usepackage{booktabs}
\usepackage{graphicx}
\usepackage{subcaption}
\usepackage{url}
\usepackage{amsmath,amssymb,mathtools}
\usepackage{siunitx}
\usepackage{enumitem}
\usepackage{float}
\usepackage[hidelinks]{hyperref}
\usepackage{balance}
\usepackage{siunitx}
\usepackage{caption}
\usepackage{tabularx,booktabs,array}
\newcolumntype{Y}{>{\raggedright\arraybackslash}X}
\usepackage{array}
\usepackage{microtype}    

\newcolumntype{Y}{>{\raggedright\arraybackslash}X}
\newcolumntype{L}[1]{>{\raggedright\arraybackslash}p{#1}}

\usepackage{enumitem}

\makeatletter
\renewcommand\subsubsection{\@startsection{subsubsection}{3}{\z@}%
  {6pt plus 2pt minus 2pt}% space above
  {3pt plus 1pt minus 1pt}% space below
  {\normalfont\normalsize\bfseries}} 
\makeatother

\usepackage[font=small,skip=4pt]{caption}

\usepackage{adjustbox}
\usepackage{array}
\usepackage{ragged2e}

\usepackage{etoolbox}

\title{Towards Senior-Robot Interaction: Reactive Robot Dog Gestures}

\author{%
  Chunyang Meng\textsuperscript{1}\
  Eduardo B. Sandoval\textsuperscript{2} \thanks{This research was partially supported by the Scientia Program PS46183-A (Dr Sandoval).}\
  Ricardo Sosa\textsuperscript{3,4}\
  Francisco Cruz\textsuperscript{1,5}\\[3pt]
  \textsuperscript{1} School of Computer Science and Engineering, UNSW Sydney, Australia\\
  \textsuperscript{2} School of Art and Design, Creative Robotics Lab, UNSW Sydney, Australia\\
  \textsuperscript{3} School of Design, University of Waikato, New Zealand\\
  \textsuperscript{4} University of Sydney, Australia\\
  \textsuperscript{5} Escuela de Ingenier\'ia, Universidad Central de Chile, Santiago, Chile\\
  Emails: \{chunyang.meng, e.sandoval, f.cruz\}@unsw.edu.au, ricardo.sosa@waikato.ac.nz
}

\date{}

\newcommand{\PaperCiteAs}{%
Cite as: Chunyang Meng, Eduardo Benitez Sandoval, Ricardo Sosa and Francisco Cruz. Towards Senior-Robot Interaction: Reactive Robot Dog Gestures. 
\emph{Proceedings of the Australasian Conference on Robotics and Automation (ACRA)}, 2025.
}

\makeatletter
\pretocmd{\@maketitle}{%
  \noindent\small \PaperCiteAs\par
  \vspace{0.5em}
  \hrule
  \vspace{-2.5em}
}{}{\PackageWarning{acra}{Failed to patch \string\@maketitle}}
\makeatother

\begin{document}
\maketitle

\begin{abstract}
As the global population ages, many seniors face the problem of loneliness. Companion robots offer a potential solution. However, current companion robots often lack advanced functionality, while task-oriented robots are not designed for social interaction, limiting their suitability and acceptance by seniors. Our work introduces a senior-oriented system for quadruped robots that allows for more intuitive user input and provides more socially expressive output.  For user input, we implemented a MediaPipe-based module for hand gesture and head movement recognition, enabling control without a remote. For output, we designed and trained robotic dog gestures using curriculum-based reinforcement learning in Isaac Gym, progressing from simple standing to three-legged balancing and leg extensions, and more. The final tests achieved over 95\% success on average in simulation, and we validated a key social gesture (the paw-lift) on a Unitree robot. Real-world tests demonstrated the feasibility and social expressiveness of this framework, while also revealing sim-to-real challenges in joint compliance, load distribution, and balance control. These contributions advance the development of practical quadruped robots as social companions for the senior and outline pathways for sim-to-real adaptation and inform future user studies.

%CHECK IF THE FONT OF ABSTRACT IS CORRECT. 
%Confirmed that this is the default style of the template.
%The abstract has been shortened and the sentence about the ablation study has been removed

\end{abstract}

\section{Introduction}

As society faces an aging population, an increasing number of older adults live alone and face challenges such as loneliness and limited social interaction. Pet therapy has been shown to alleviate these issues by providing companionship and emotional support \cite{cryer_pawsitive_2021}. However, many seniors are unable to care for real animals due to physical or cognitive limitations. Companion robots may offer a potential alternative, yet many current offerings are designed as passive, plush companions that lack the mobility and practical functionality to assist with daily tasks.

Quadruped robots have demonstrated strong navigation and task execution capabilities\cite{majithia_design_2024}. However, their performance in  Human-Robot Interaction (HRI) remains relatively underexplored. Specifically, their movements can appear unnatural and they often lack rich interactive capabilities, which can make users, especially older adults, feel uncomfortable. In particular, there is a lack of design frameworks that combine intuitive user control with socially meaningful robot behaviours tailored for older adults\cite{schneider_coaching_2023}. In this work, we aim to provide social communication skills to service robots such as robot dogs. We believe that this capability will have a positive impact on the future of Senior-Robot Interaction (SRI).

Based on these insights, we propose an interaction framework for older adults that combines a simple interface to capture a series of human gestures and trigger expressive gestures by the robot dog. 
Our main contributions are as follows: 
\begin{itemize}
    \item Our framework allows users to intuitively guide the robot using hand and head movements recognised by MediaPipe\cite{zafar_real-time_2024}\cite{shin_non-verbal_2025}.
    % ?
    \item In addition to basic locomotion, the robot can perform communicative behaviours such as attracting attention, pointing to indicate direction, and waving to invite a handshake.
    \item We designed a staged curriculum that progresses from stable standing to three-legged balancing and forward extension, combined with reward-shaping mechanisms to achieve load redistribution and motion consistency.
    \item We implemented a sim-to-real pipeline under low-level control and conducted an ablation study to evaluate the role of curriculum learning. We validated our method through simulations in Isaac Gym and deployment on a Unitree Go1 robot, demonstrating the advantages of CL-based RL and the challenges of sim-to-real transfer.
\end{itemize}

\section{Related Work}

\subsection{HRI in Senior Care}
With the global aging population, companion robots are seen as an important tool to reducing loneliness and improve the well-being of seniors. A study has shown that animal-like forms and gentle, predictable behaviours of robots such as PARO can improve mood and encourage sustained use \cite{inoue_exploring_2021}. Other studies indicate that older adults' acceptance of robots largely depends on dignity and ease of use \cite{coghlan_dignity_2021}\cite{costanzo_new_2024}, suggesting that seniors are unlikely to adopt robots that behave unpredictably or are difficult to control. Some research has also begun adapting quadruped robots for elder care, for example by adding safety mechanisms and dog-like cues to the Boston Dynamics Spot robot \cite{cavazos_robotic_2025}.

Despite these advances, a gap remains between comforting and functional robots. On the one hand, comforting robots such as PARO, are often designed as plush companions, lacking mobility and functionality. On the other hand, quadruped robots are typically designed for functional tasks, lacking social expressiveness and interactive capabilities with users. Our work seeks to bridge this gap by enabling users to control the robot intuitively through hand gestures or head movement, and by enabling the quadruped robot to perform gestures that are both socially expressive and functionally practical. Such integration of social expressiveness and practicality could offer benefits for older users: the robot could be more emotionally acceptable and also assist with daily tasks in a more intuitive way. In other words, social robots could simultaneously enhance emotional well-being and provide tangible functional support, making them more functional and valuable companions in daily life. This aligns with the vision of service robots equipped with social interfaces \cite{sandoval_industrial_nodate}, where interaction design informs use cases and enhances both usability and dignity in elder care.
For this implementation, we mainly used two large blocks of development: a) Gesture Recognition in HRI, and b) Curriculum Learning based Reinforcement Learning for training the social gestures of the robot. 

%% What are the benefits for the users in having more social utilitarian robots around them?

%% YOu can cite Eduardo's patent about social interfaces in service robtos. Mexican patent. Service Robot with Social Interface (Registered Industrial Design) \cite{sandoval_industrial_nodate}
%Inventors
%Eduardo B. Sandoval, Hector Castillo, Samuel Tellez
%Publication date
%2019/1
%Patent office
%MX
%Patent number
%MX/f/2017/002789
%Application number
%MX/f/2017/002789
%https://unsworks.unsw.edu.au/entities/publication/3e6a044f-bab4-4fc4-88a0-5d759e445bd4/full
%For this implementation we used mainly two large blocks of development: a) Gesture Recognition in HRI, and b) Curriculum Learning and Reinforment Learning for training the social gestures of the robot.

%Answered the question "benefits for the users in having more social utilitarian robots around them"; cited the patent; Add "For this implementation ..." part

\subsection{Gesture Recognition in HRI}

For some users, especially older adults, it can be difficult to use handheld devices or complex software\cite{czaja_current_2021}. Therefore, gesture-based control offers a more intuitive way to control robots. Lightweight frameworks like MediaPipe provide real-time hand and face tracking on low-cost hardware \cite{lugaresi_mediapipe_2019}\cite{zhang_mediapipe_2020}, which makes them well-suited for accessible HRI. Previous studies demonstrate that gesture control improves user satisfaction and task efficiency when interacting with quadruped robots \cite{zafar_real-time_2024}\cite{shin_non-verbal_2025}, and advanced pipelines have integrated 3D sensing or sequence models to map dynamic gestures to robot motion \cite{xie_humanrobot_2025}. These results confirm that gesture control provides a reliable and comfortable interaction solution for older adults. Based on these findings, our system integrates hand gesture and head movement recognition into the control scheme of a quadruped robot, enabling users to operate the robot through intuitive gestures or simple head movements. The reliability and performance of our gesture recognition system, including latency and accuracy under different lighting conditions, are evaluated and presented in Section~\ref{sec:user-control}.

%Where do you talk about the reliability and results of the gesture recognition?
%presented in Section~\ref{Gesture-Based User Control}.

\subsection{Curriculum Learning Based Reinforcement Learning}

Curriculum Learning (CL) is a training strategy that begins with simpler tasks and gradually increases complexity \cite{bengio_curriculum_2009}. A previous work has shown its benefits in improving convergence and generalization, particularly in reinforcement learning (RL) where rewards are sparse or objectives are highly complex \cite{soviany_curriculum_2022}. In robotics, CL has been successfully applied to motion planning \cite{zhou_robotic_2021}, stepping-stone walking \cite{xie_allsteps_2020}, agile locomotion \cite{hwangbo_learning_2019}, and multi-task jumping control \cite{li_robust_2023}. Atanassov et al. \cite{atanassov_curriculum-based_2025} specifically applied a staged CL framework to a Unitree Go1 robot, demonstrating efficient training of precise jumping behaviours.  Margolis and Agrawal \cite{margolis_walk_2022} proposed the Walk These Ways controller for the Go1, which encodes a multiplicity of locomotion strategies to enhance robustness and generalization, providing valuable insights into training and deployment on real hardware. A study shows that curriculum design and structured locomotion policies help bridge the gap between simulation and real-world deployment \cite{tang_deep_2025}. Building on the existing literature, we adopt a CL-based RL approach to train the quadruped robot to perform socially expressive gestures. The training curriculum progresses from four-leg stability to the more advanced skill of three-leg balancing, which then serves as the foundation for learning various socially expressive gestures.

\section{Method}

This section discusses the methods and technical details of the design of our social gestures implemented in the robot dog. We provide a system overview and a detailed description of the gesture recognition for high-level control and the complex social gestures for low-level control.

%Check about the low level gestures and the Complex Social Gestures
% This section has been reorganized.

\subsection{System Overview}
\begin{figure}[h]
  \centering
  \includegraphics[width=0.95\columnwidth]{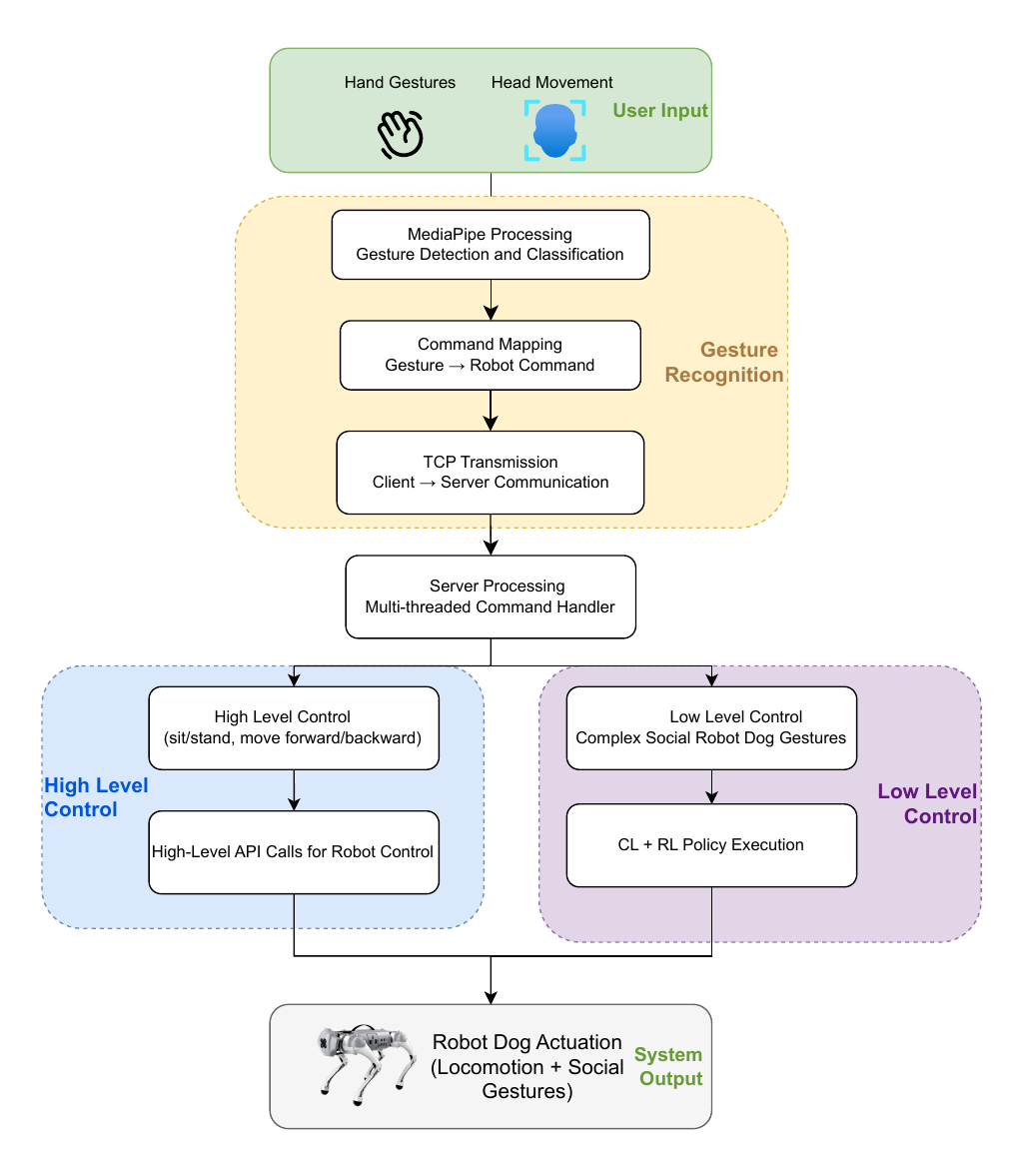}
  \captionsetup{font=small, justification=raggedright, singlelinecheck=false}
  \caption{System architecture: A MediaPipe-based module processes user gestures into commands, which are sent via TCP to a server. The robot runs a high-level controller for posture and simple locomotion via API calls or a low-level controller that executes complex social gestures using curriculum-learning reinforcement-learning (CL-RL) policies.}
  \label{fig:system}
\end{figure}

Figure~\ref{fig:system} illustrates the overall architecture. 
User gestures (hand or head movements) are captured by a camera and processed by a MediaPipe-based recognition module that maps them into discrete robot commands. 
These commands are transmitted through a TCP bridge to a server, which routes them to either High-Level Control or Low-Level Control. 
High-Level Control uses API calls to trigger basic motions such as sitting, standing, or locomotion, while Low-Level Control executes curriculum-trained RL policies for complex social gestures, such as paw lift, pointing, waving. 
Both control pathways converge in the robot actuation stage, where commands are executed on the quadruped platform via the Unitree SDK.

\vspace{-1em}

\subsection{High-Level Control Path}

High-level control relies on API calls to trigger predefined robot actions. We use MediaPipe to recognise hand and head gestures, which are mapped to commands such as move forward, move backward, sit, and stand etc.

\subsubsection{Gesture-to-Command Mapping}
MediaPipe Hands and Face Mesh detect seven gestures. 
Each gesture maps to a discrete command sent via TCP to the Unitree controller (Table~\ref{tab:gesture_rules}).

Distances are normalized by the hand bounding-box width; 
finger/joint angles are PIP(proximal interphalangeal)/MCP(metacarpophalangeal) internal angles (degrees) from MediaPipe; 
temporal windows are $\sim$400~ms with exponential smoothing 
($\alpha \approx 0.6$); 
nod/shake thresholds use nose\_threshold $\approx 0.015$ 
and jaw\_threshold $\approx 0.02$.

\begin{table}[h]
\centering
\captionsetup{font=small}
\caption{Gesture-to-command mapping and detection rules}
\label{tab:gesture_rules}
\setlength{\tabcolsep}{4pt}
\renewcommand{\arraystretch}{1.1}
{\footnotesize
\begin{tabularx}{\columnwidth}{@{}l l Y l@{}}
\toprule
\textbf{Input} & \textbf{Gesture} & \textbf{Rule} & \textbf{Command} \\
\midrule
Hand & Open Palm  & Four fingers extended (tips above PIP); thumb extended; adjacent-finger spacing $>0.03$ in $x$ & \texttt{MOVE\_FWD} \\
Hand & Fist       & All fingers flexed (angles $<160^\circ$); thumb not extended; fingertips near palm center ($<0.15$) & \texttt{MOVE\_BWD} \\
Hand & Thumb Up   & Thumb tip $>$ MCP by $0.05$; others flexed; thumb is highest fingertip & \texttt{SPEED\_UP} \\
Hand & Thumb Down & Thumb tip $<$ MCP by $0.05$; others flexed; thumb is lowest fingertip & \texttt{SPEED\_DOWN} \\
Hand & Pointing Up      & Only index extended with tip $>$ MCP by $0.05$; others flexed; index is highest fingertip & \texttt{STOP} \\
Head & Nod        & $\ge 2$ pitch sign-changes with amplitude $>\textit{nose\_threshold}$ within the window & \texttt{STAND} \\
Head & Shake      & $\ge 2$ yaw sign-changes with amplitude $>\textit{jaw\_threshold}$ within the window & \texttt{SIT} \\
\bottomrule
\end{tabularx}
}
\end{table}
\vspace{-1em}

\subsubsection{Robustness and Safety}
To ensure robust recognition and safe actuation, the pipeline incorporates three key mechanisms:
\begin{itemize}[noitemsep, topsep=0pt, leftmargin=*]
    \item \textbf{Noise filtering:} short-window temporal smoothing, multi-condition landmark checks, and gesture-specific thresholds (e.g., nod vs.\ shake) to reduce spurious detections.
    \item \textbf{Duplicate prevention:} gestures are only transmitted on change events, with a \SI{1}{\second} cooldown. The server maintains a thread-safe record of the last gesture and prevents TCP packet concatenation issues.
    \item \textbf{Safe motion execution:} any new motion immediately halts the previous one. Continuous actions run in isolated loops, while posture commands (e.g., stand/sit) are lock-protected to avoid conflicts.
\end{itemize}

\subsection{Low-Level Control Path}

We used curriculum learning based reinforcement learning to train five complex gestures, as shown below.

\subsubsection{Design of Complex Social Robot Dog Gestures}
We design five complex gestures (G1--G5) that map learned motion primitives to social and functional intents. These gestures are inspired by real canine body language\cite{worsley_cross-species_2018} and human communicative gesture\cite{robinson_robotic_2023}, and are adapted to quadruped robots to support attention, guidance, anticipation, and playful interaction.
For the design of complex social robot dog gestures, see Table \ref{tab:gesture_design} for further details.

%MAKE THE TABLE HORIZONTAL
%MAKE Every of the three stages a sub-subsection

\begin{table}[h]
\centering
\captionsetup{font=small}
\caption{Gestures for Social Interaction: Design and Description}
\label{tab:gesture_design}
\renewcommand{\arraystretch}{1.12}
\setlength{\tabcolsep}{4pt}
{\small
\begin{tabularx}{\linewidth}{@{}L{0.18\linewidth}Y@{}} % 
\textbf{Gesture} & \textbf{Description} \\
\midrule
G1 Attract Attention & Robot stands on three legs with the front-left leg slightly raised (paw lift). Intended to attract user attention or invite interaction. Inspired by dogs’ paw-lifting to signal curiosity and anticipation. \\
G2 Go Left & Robot raises and extends the front-left leg forward (pointing). Guides the user toward the front-left direction. Visual directional gestures are intuitive and accessible for older adults. \\
G3 Go Right & Robot raises and extends the front-right leg forward (pointing). Guides the user toward the front-right direction, complementing G2 to enable bidirectional guidance. \\
G4 Waiting & Robot rhythmically lifts and lowers its front-left leg in place. Conveys a “waiting” state more engagingly than static postures; dynamic motion sustains user focus. \\
G5 Waving & Robot extends the front-left leg forward and swings it gently (paw offer). Invites handshake or playful interaction; encourages engagement through familiar pet-like behaviour. \\
\bottomrule
\end{tabularx}
}
\end{table}
\vspace{-1em}

\subsubsection{MDP Formulation}

We model each gesture generation task as a Markov Decision Process (MDP), defined by the tuple $(\mathcal{S}, \mathcal{A}, P, R, \gamma)$, as follow:

\begin{description}[style=unboxed, leftmargin=0pt, itemsep=2pt, topsep=3pt] 
\item[State Space ($\mathcal{S}$).]
The state $s_t \in \mathcal{S}$ includes the robot's base pose (position $p_t$ and orientation as a quaternion), base linear and angular velocities $(v_t, \omega_t)$, joint positions and velocities $(q_t, \dot{q}_t) \in \mathbb{R}^{12}$, and a history of the last actions.

\item[Action Space ($\mathcal{A}$).]
The action $a_t \in \mathcal{A} \subset \mathbb{R}^{12}$ is a vector of target joint positions sent to the low-level PD controllers in the simulator.

\item[Transition ($P$).]
The state transition dynamics $s_{t+1} \sim P(\cdot|s_t, a_t)$ are governed by the physics engine (Isaac Gym), which simulates contact dynamics, gravity, and domain randomization effects such as variations in mass, friction, and motor strength.
\end{description}

%\paragraph{Curriculum-Based Reward Function ($R$).}
\subsubsection{Curriculum-Based Reward Design}
To train the complex gestures, we design the reward function $R$ based on a formal three-stage curriculum. Each stage builds upon the last by introducing new reward components, progressively increasing the task difficulty. This approach avoids instabilities that arise from training the final gesture directly.

\begin{description}[style=unboxed, leftmargin=0pt, itemsep=4pt, topsep=5pt] 

\item[Notation and operators.]\label{par:notation}
$\|\cdot\|_1$ and $\|\cdot\|$ denote the L1 and L2 norms, respectively;
$[x]_+ \triangleq \max(0,x)$;
$\sigma(x) \triangleq \tfrac{1}{1+e^{-x}}$;
$\operatorname{dist}(x,S)$ is the Euclidean distance from point $x$ to a set/line $S$.
$\mathrm{CoP}$ is the center of pressure; $L_{\mathrm{anchor}}$ is the closest edge of the current support triangle.
Leg labels: FL (front-left), FR (front-right), RL (rear-left), RR (rear-right).
$F_{z,i}$ is the vertical ground-reaction force at foot $i$; $F_{\min}$ is the minimum support-force threshold; $F_{\mathrm{thresh}}$ is the near-zero contact threshold.
$\alpha \in \mathbb{R}^3$ are the normalized vertical-force ratios on the three support feet with target $\alpha^{\star}$;
$\phi_{\mathrm{FL}}$ is the FL force fraction with target $\phi^{\star}$.
$z$ is the FL foot world height; $z_{\mathrm{rel}}$ is the foot height relative to ground; $z^{\star}$ is the target height.
$p_{xy}\!\in\!\mathbb{R}^2$ is the FL foot body-frame $(x,y)$ position; $p_{xy}^{\star}$ is the forward target position.
$t_{\mathrm{air}}$ is the cumulative off-ground time; $t^{\star}$ is the target air-time; $\Delta t$ is the control step.
Let \textit{support} denote the set of feet in contact at a given time.
All gains $k_{\{\cdot\}}\!>\!0$ are scalars, and $\delta\!>\!0$ is a safety margin.

%\vspace{1ex} %
%\noindent\textbf{Stage 1: Stable Four-Leg Standing} %MAke it a sub sub section all of the three stages
\item[Stage 1: Stable Four-Leg Standing.]
%\begin{description}[font=\normalfont]
    \textbf{Goal:} The initial stage aims to learn a preparatory stance for the subsequent leg-lift. The goal is not merely stable standing, but to proactively redistribute the robot's weight onto a stable tripod base (FR, RL, RR) while significantly unloading the front-left (FL) leg to a small target force fraction, all while keeping all four feet on the ground.

    \textbf{Reward Function ($R_A$):} The reward consists of the base stability terms ($R_{\text{base}}$), augmented with crucial task-specific rewards to achieve the preparatory tripod stance. The main components are:
    \textit{Tripod Force Distribution} ($r_{\text{distribute}}$), which encourages the vertical force ratios $\alpha$ on the three support feet to match a target $\alpha^{\star}$;
    \[
      r_{\text{distribute}} = \exp\!\big(-k_d \,\|\alpha - \alpha^{\star}\|_1\big)
    \]
    \textit{CoP Stability} ($r_{\text{cop}}$), which ensures the Center of Pressure is safely contained within the support triangle;
    \[
      r_{\text{cop}} = \exp\!\big(-k_c\,\big[\operatorname{dist}(\mathrm{CoP},L_{\mathrm{anchor}})-\delta\big]_+\big)
    \]
    \textit{FL Unloading} ($r_{\text{unload}}$), which rewards reducing the force fraction of the FL leg, $\phi_{\mathrm{FL}}$, towards a small target value $\phi^{\star}$;
    \[
      r_{\text{unload}} = \exp\!\big(-k_u\, [\phi_{\mathrm{FL}} - \phi^{\star}]_+\big)
    \]
    This proactive unloading strategy serves as a critical warm-start, making the subsequent transition to a full leg-lift in Stage 2 substantially more stable and easier to learn.

    \textbf{Success Criterion:} The policy is considered successful when it consistently achieves the target FL leg force fraction ($\phi_{\mathrm{FL}} \approx \phi^{\star}$) while maintaining all stability metrics (CoP, orientation, height) for the tripod support structure.
%\end{description}

\item[Stage 2: Three-Leg Balance with Leg Lift.]
%\begin{description}[font=\normalfont]
    \textbf{Goal:} Building on the stable standing policy, the robot learns to lift one front leg (FL) while maintaining stable balance on the remaining three legs (the tripod base).

    \textbf{Reward Function ($R_B$):} The reward function is augmented with components grouped into two categories.
    \textit{Tripod Stability Rewards ($R_{\text{tripod}}$)} ensure the robot does not fall over, and include $r_{\text{distribute}}$ and $r_{\text{cop}}$ as defined previously, along with \textit{Minimum Support} ($r_{\text{support}}$), which penalizes any support foot's force dropping below a threshold $F_{\min}$:
    \[
      r_{\text{support}} = \prod_{i \in \mathrm{support}} \sigma\!\big(k_s\,(F_{z,i} - F_{\min})\big)
    \]
    \textit{FL Leg Lift Rewards ($R_{\text{lift}}$)} guide the desired gesture. These include \textit{No Contact} ($r_{\text{no-contact}}$), which rewards near-zero force on the lifted foot:
    \[
      r_{\text{no-contact}} = \sigma\!\big(k_n\,(F_{\mathrm{thresh}} - F_{z,\mathrm{FL}})\big)
    \]
    \textit{Clearance} ($r_{\text{clearance}}$), which rewards the relative vertical height $z_{\mathrm{rel}}$ of the foot above a target $z^{\star}$:
    \[
      r_{\text{clearance}} = \big[z_{\mathrm{rel}} - z^{\star}\big]_+
    \]
    and \textit{Air Time} ($r_{\text{air-time}}$), which rewards the cumulative time $t_{\mathrm{air}}$ the foot remains off the ground:
    \[
      r_{\text{air-time}} = \min\!\big(1,\; [\,t_{\mathrm{air}} - t^{\star}\,]_+ \,/\, \Delta t\big)
    \]

    \textbf{Success Criterion:} Success is defined by a high leg-lift success rate ($>95\%$), stable base height control during the lift, and effective load redistribution onto the three supporting legs.
%\end{description}

\item[Stage 3: Three-Leg Balance with Forward Extension.]
%\begin{description}[font=\normalfont]

    \textbf{Goal:} The final stage refines the leg-lift policy to extend the lifted leg towards a forward target, creating a ``pointing'' gesture.

    \textbf{Reward Function ($R_C$):} This stage modifies the lift rewards for precision and adds a crucial extension reward.
    The \textit{Modified Lift Reward} is \textit{Target Height} ($r'_{\text{clearance}}$), which rewards the foot's world height $z$ matching a specific target $z^{\star}$:
    \[
      r'_{\text{clearance}} = \exp\!\big(-k_h \,\|z - z^{\star}\|^2\big)
    \]
    The primary \textit{FL Leg Extension Reward} is \textit{Forward XY} ($r_{\text{extend}}$), which rewards the foot's body-frame XY position $p_{xy}$ matching a forward target $p^{\star}_{xy}$ (nominal stance plus a forward offset along the body $x$-axis):
    \[
      r_{\text{extend}} = \exp\!\big(-k_p \,\|p_{xy} - p^{\star}_{xy}\|^2\big)
    \]

    \textbf{Success Criterion:} The policy must reach the forward target with the foot (e.g., small error $<0.1$ rad in relevant joints) while satisfying all stability gates.
%\end{description}

\item[Gating and Regularization.]
Most task-specific rewards (e.g., $r_{\text{clearance}}$, $r_{\text{extend}}$) are multiplied by a \textit{stability gate} $g_{\text{stable}} \in [0, 1]$. This gate function combines orientation stability and CoP stability, ensuring that the robot only receives rewards for performing the desired gesture when its tripod base is stable. Additional regularization terms penalize high joint torques and deviations from a default joint configuration to promote energy efficiency and natural postures.
\end{description}

\subsubsection{Reinforcement Learning Setup}
Each task is trained with Proximal Policy Optimization (PPO) using the rsl\_rl library. 
Policies are represented by a three-layer MLP ([256, 128, 64], ELU). 
We adopt a clipping range of $0.15$, discount factor $\gamma=0.998$ (four-leg) 
or $0.995$ (three-leg), GAE parameter $\lambda=0.95$, and an entropy coefficient 
between $0.01$–$0.015$. Training uses 64 parallel environments with 
$\approx 3{,}000$–$4{,}000$ steps per iteration, and later tasks initialize 
from earlier checkpoints (curriculum transfer). 
We employ domain randomization and random state initialization to further improve robustness.

\subsection{Implementation (Safety and Control)}
Safety gates terminate episodes when roll/pitch exceed limits, base height 
leaves bounds, or self-collision occurs. 
Posture commands (stand/sit) are mutually exclusive, and 
a ``new motion stops old motion'' rule avoids conflicts. 
Slip and excessive torque are penalized to encourage stable execution.

\subsection{Experimental Scenario}

\subsubsection{User-to-Robot Control Setup}
We evaluate the gesture recognition and command transmission in a controlled test conducted by a single operator. The evaluation is performed under varied camera viewpoints and lighting conditions to assess robustness. MediaPipe runs in real time on a laptop camera, mapping hand and head gestures to robot commands. Key metrics include recognition accuracy and end-to-end gesture recognition latency (camera-to-command).

\subsubsection{Robot-to-Human Expression Setup}
To evaluate the expressive gesture policies on hardware, trained networks 
are exported and deployed to a Unitree Go1 robot using the official SDK 
at \SI{20}{Hz}. Actions are issued as joint-position deltas from a nominal 
pose, smoothed and clamped within safe bounds. A settle–active–return 
schedule is used to ensure safe execution. Deployment takes place on flat 
indoor terrain under conservative operating limits. Torque sensing and 
latency compensation are not included in the current implementation. 
Evaluation focuses on stability, gesture amplitude, and consistency with 
simulation rollouts across the three curriculum stages 
(four-leg standing, leg lift, forward extension).

\section{Results}

\subsection{Gesture-Based User Control}\label{sec:user-control}
\begin{figure}[h]
  \centering
  \subfloat[Correct detection: open palm under dim light]{%
    \includegraphics[width=0.45\columnwidth]{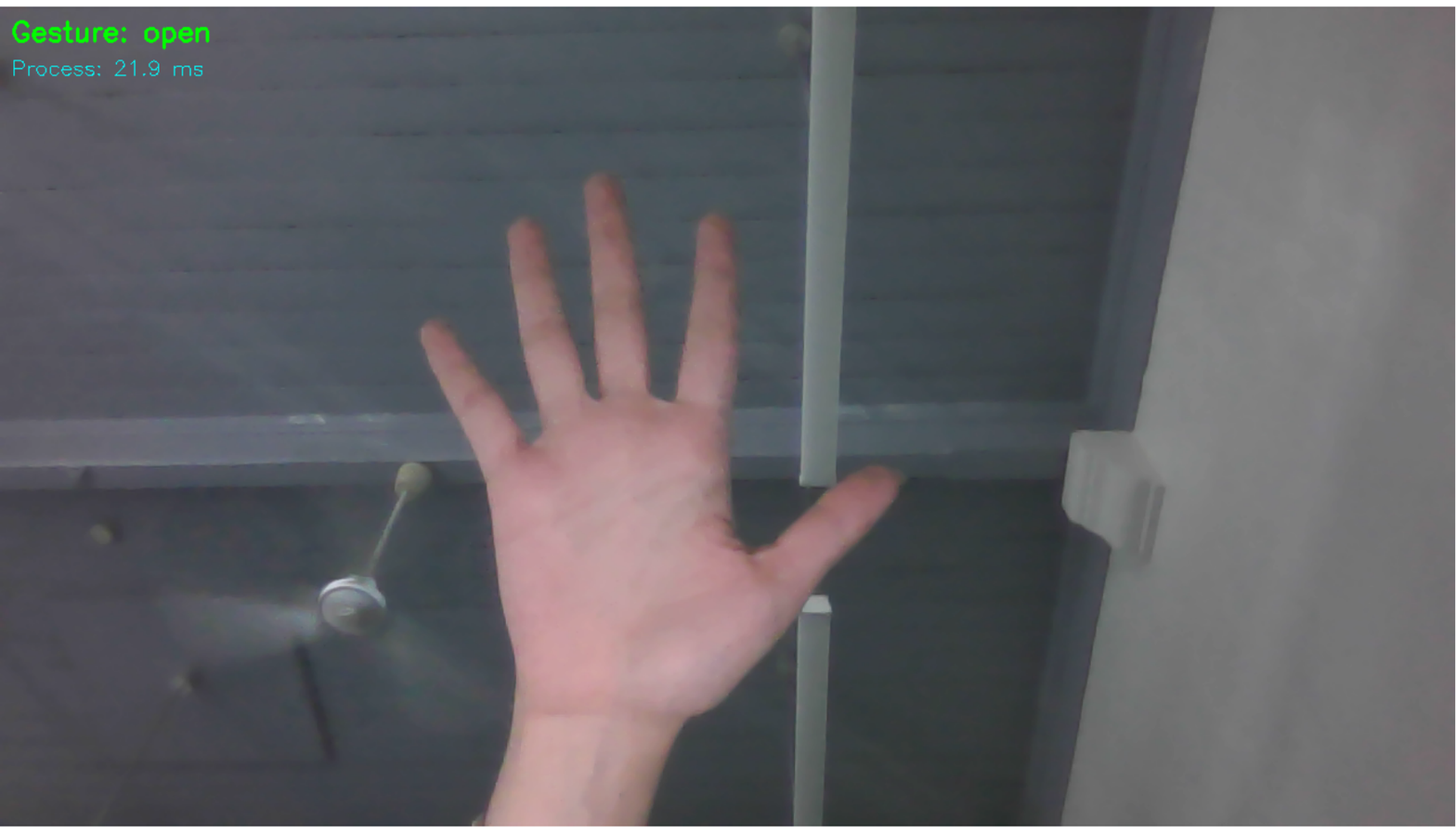}}
  \hfill
  \subfloat[Failure case: pointing-up misclassified as a fist]{%
    \includegraphics[width=0.45\columnwidth]{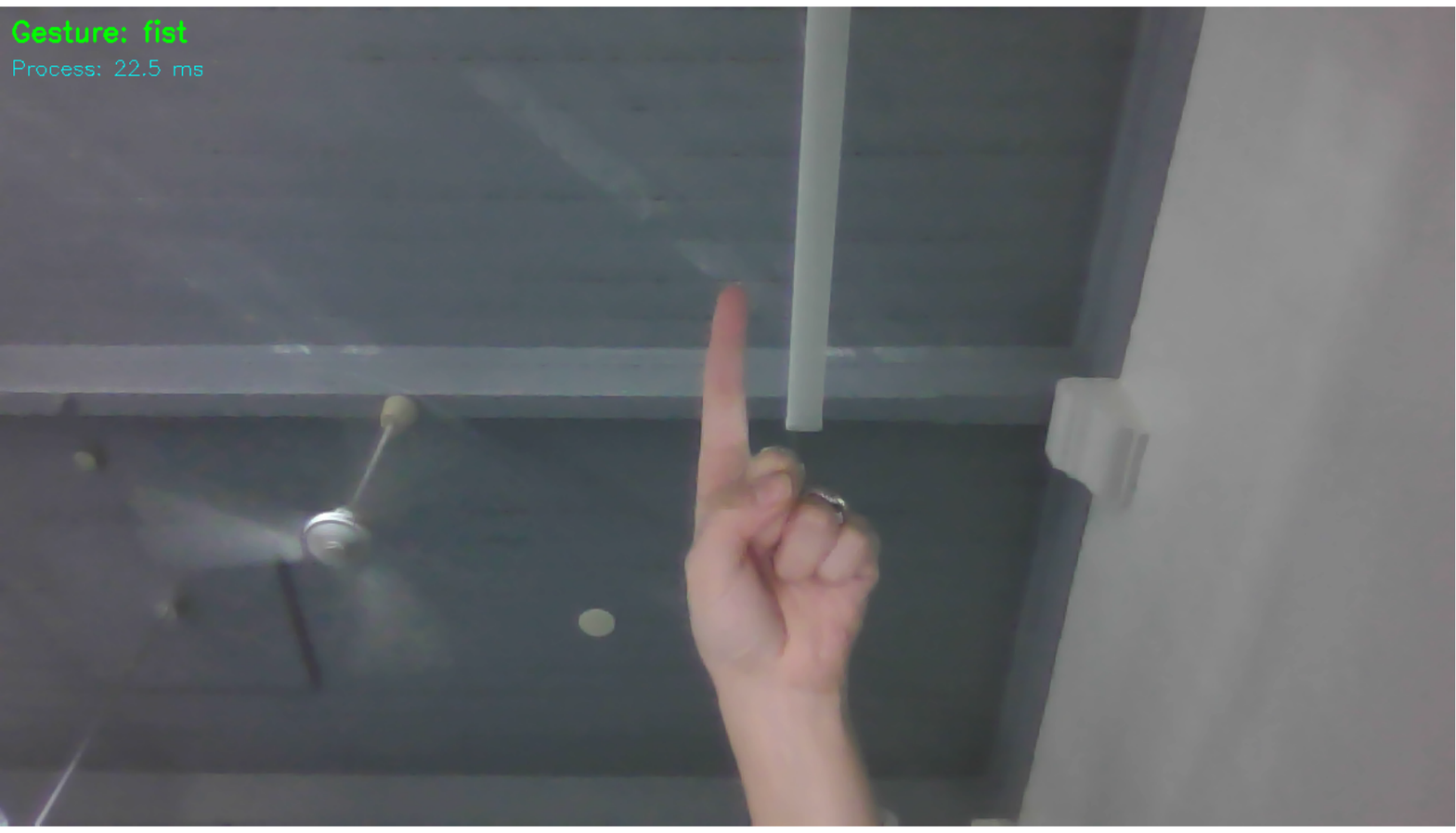}}
  \captionsetup{font=small}
  \caption{Examples of gesture detection under dim lighting. 
  (a) Correct recognition of an open palm. 
  (b) Misclassification of pointing up as a fist.}
  \label{fig:gesture_examples}
\end{figure}

We evaluated the MediaPipe-based perception module in controlled tests conducted by the author under varied camera viewpoints and lighting conditions. 
The predefined gesture set covered open palm/fist/thumb up/thumb down/pointing up for the hand channel and nod/shake for the head channel; temporal smoothing with duplicate filtering reduced spurious repeats. 
Examples are shown in Fig.~\ref{fig:gesture_examples}. Integration with the Unitree SDK confirmed end-to-end operation without hand-held devices. 
Head movements also provide an alternative input channel that may be preferable when hand use is limited.

To assess responsiveness and reliability, we measured end-to-end gesture-recognition latency (camera to command) and offline classification accuracy under bright and dim conditions. 
Results are summarised in Table~\ref{tab:gesture_latency_accuracy}. 
For \textbf{hand gestures}, the system achieved consistently low delays of about \SIrange{19}{22}{\milli\second}, with minimal variation across conditions. 
Recognition accuracy was high in bright light (96\%, 24/25) but dropped to 84\% (21/25) in dim light, where most errors resembled Fig.~\ref{fig:gesture_examples}(b). 
For \textbf{head gestures}, latency was similarly low (\SIrange{8}{9}{\milli\second}). 
By contrast with hand gestures, accuracy remained perfect under both lighting conditions (100\%).
These results suggest that hand gestures are intuitive for primary control but need improved robustness in low light, while head gestures, though less diverse, provide a fast and highly reliable auxiliary channel.

\begin{table}[h]
\centering
\captionsetup{font=small}
\caption{Gesture recognition under different lighting conditions.
Latency in milliseconds; range indicates the minimum–maximum observed latency. 
Accuracy over 25 hand-gesture trials and 10 head-motion trials (single operator).}

\label{tab:gesture_latency_accuracy}
\renewcommand{\arraystretch}{1.2}
\setlength{\tabcolsep}{4pt}
\begin{tabular}{lcccc}
\toprule
\textbf{Input} & \textbf{Lighting} & \textbf{Latency} & \textbf{Range} & \textbf{Accuracy} \\
\midrule
Hand  & Bright & 22.0 & 16.6--27.4 & 96\% (24/25) \\
Hand  & Dim    & 19.8 & 16.7--36.7 & 84\% (21/25) \\
Head  & Bright & 9.4  & 3.4--10.5  & 100\% (10/10) \\
Head  & Dim    & 8.3  & 3.3--10.7  & 100\% (10/10) \\
\bottomrule
\end{tabular}
\end{table}
\vspace{-1em}

\paragraph{Ethics note}
These were engineering tests with a single operator; no participants were recruited and no identifiable personal data were collected or stored. 
A formal user study is outside the scope of this paper.

\subsection{Robot Gesture in Simulation}

\begin{figure*}[h]
\centering
\captionsetup{font=small}
\begin{subfigure}{0.32\linewidth}
  \includegraphics[width=\linewidth]{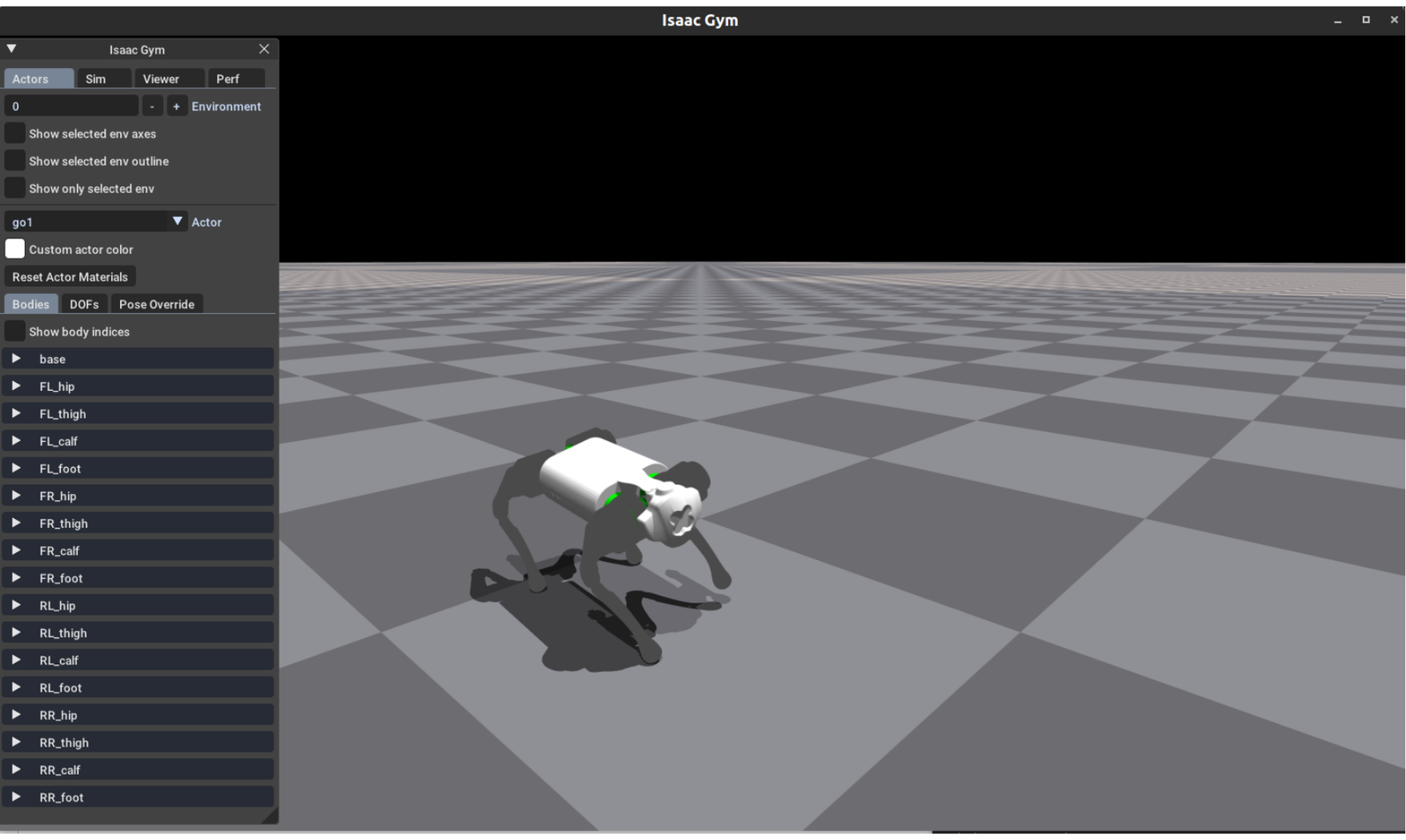}
  \caption{FL lift Gesture}
  \label{fig:robot}
\end{subfigure}
\hfill
\begin{subfigure}{0.32\linewidth}
  \includegraphics[width=\linewidth]{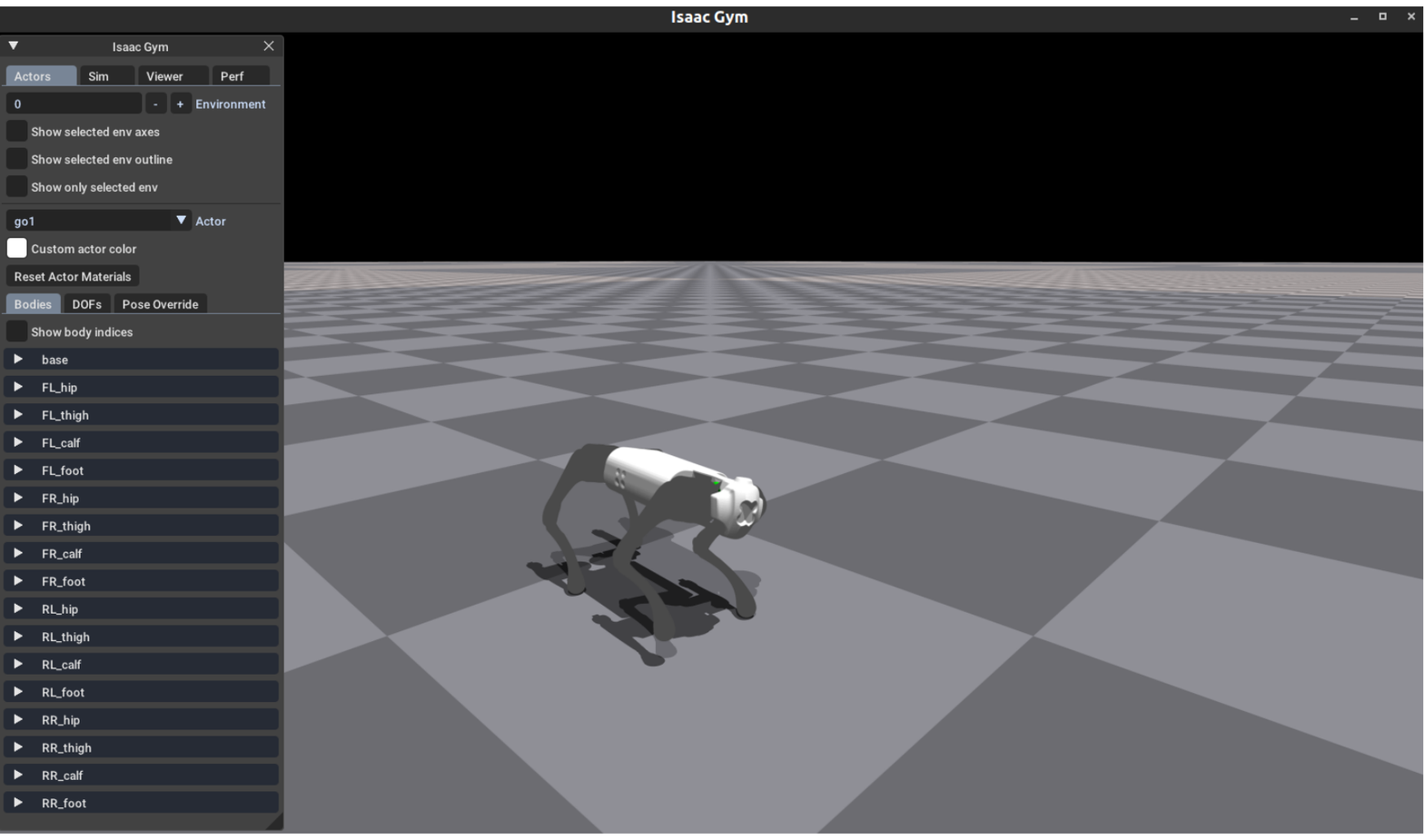}
  \caption{FR lift Gesture}
  \label{fig:sim}
\end{subfigure}
\hfill
\begin{subfigure}{0.32\linewidth}
  \includegraphics[width=\linewidth]{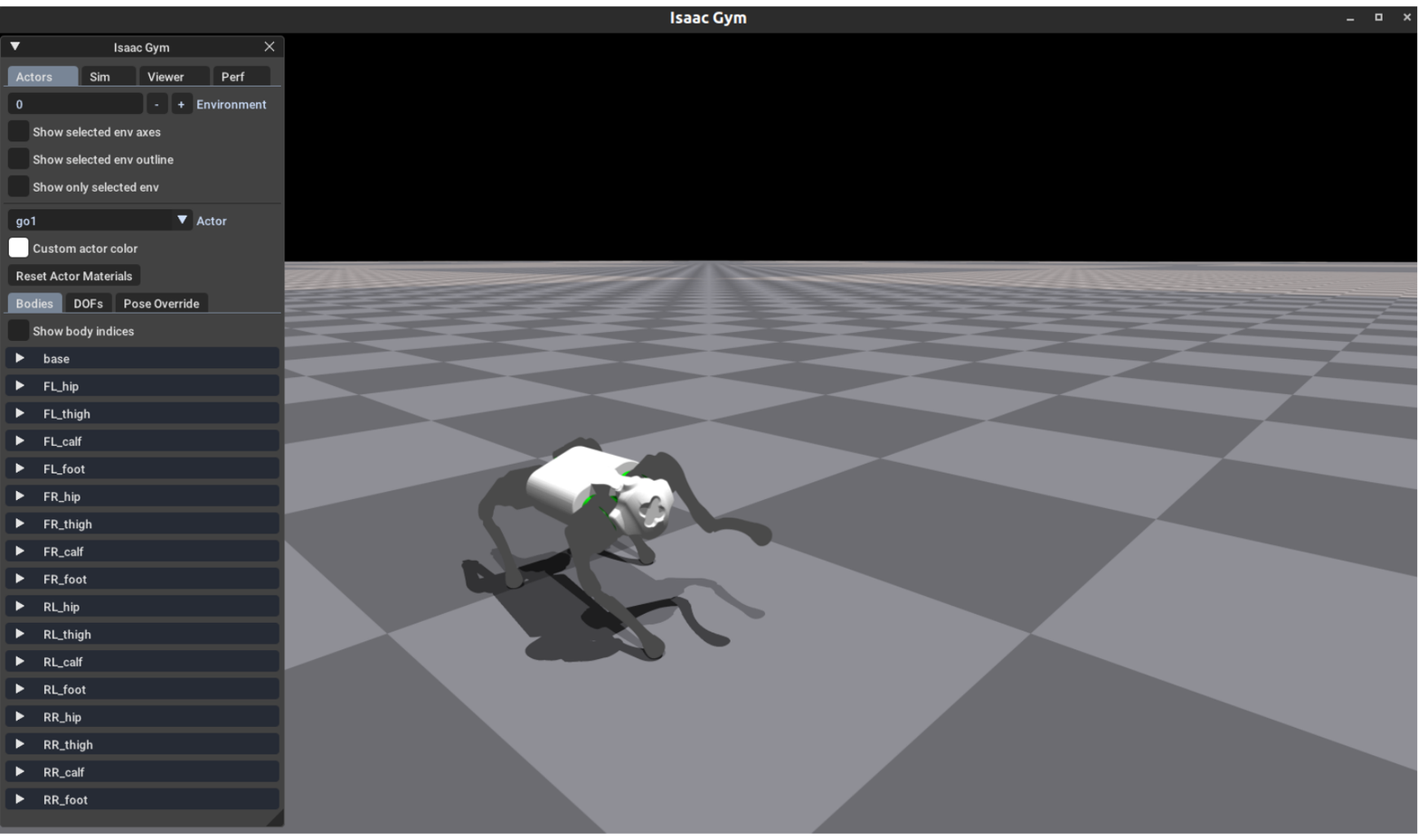}
  \caption{FL extend Gesture}
  \label{fig:gesture}
\end{subfigure}
\caption{Overview of three gesture primitives executed by the quadruped robot.}
\label{fig:overview}
\end{figure*}

\textbf{Basic standing and leg-lift:} The reinforcement learning framework successfully trained the robot to perform socially expressive gestures through a three-stage curriculum. Figure~\ref{fig:overview} summarises the three gestures: FL lift (a), FR lift (b), and FL extend (c).

Four-leg standing: The robot maintained a stable posture with a mean body height of 0.306 m (target 0.300 m, error 0.006 m), 98.8\% contact time, and intentional rear-leg load bias (~57\%).

Three-leg standing with leg lift: The policy achieved a 99.0\% leg-lift success rate, stable height control (mean 0.270 m, error 0.009 m), and effective weight redistribution across the three support legs.

Three-leg standing with leg extension: The robot extended the lifted leg while maintaining balance, with a 99.1\% leg-lift success rate, 98.8\% three-leg contact time, and mean height error of 0.009 m. Rewards increased compared to simpler standing tasks, confirming successful progression through the curriculum.

\begin{figure}[h]
\centering
\captionsetup{font=small}
\includegraphics[width=0.95\columnwidth]{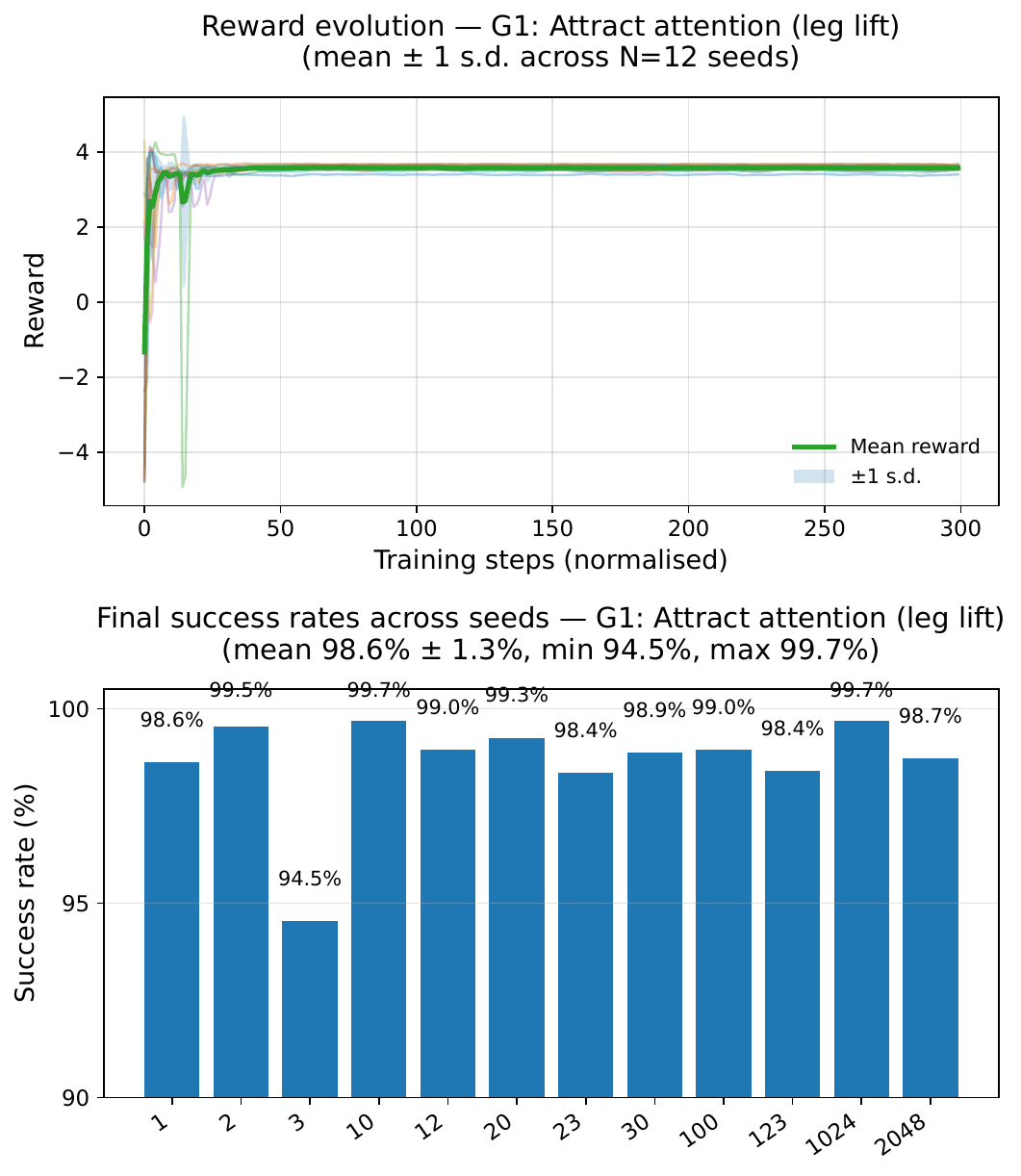}
\caption{Training performance across random seeds for Stage~2 (leg lift). 
Top: reward evolution (mean $\pm$ 1 s.d., $N{=}12$). 
Bottom: final success rates per seed (range 94.5\%–99.7\%, mean $98.6\%\pm1.3\%$). 
Note: the x-axis shows normalized training progress [0–1], while reward values are unnormalized (raw scale).}
\label{fig:combined_results}
\end{figure}

\textbf{Generalisation across seeds:} 
To evaluate robustness, we trained the Stage~2 (leg lift) policy with $N{=}12$ different random seeds. As shown in Fig.~\ref{fig:combined_results}, the final success rates were consistently high, ranging from 94.5\% to 99.7\% (mean $98.6\%\pm1.3\%$). These results confirm that the trained policies are stable and generalizable within simulation. 

\vspace{-1.5em}
\setlength{\textfloatsep}{5pt}
\setlength{\floatsep}{5pt}

%----figure here

\textbf{Composite gestures (G4, G5):} 
In addition to the basic standing and leg-extension gestures, we also designed two more expressive composite gestures by building on the same three-leg control framework. In simple terms, G4 is a 'waiting' gesture achieved by rhythmic paw lifts, and G5 is a 'waving' gesture realized by extending and swinging the lifted paw forward. These composite gestures illustrate how complex, socially meaningful motions can be composed from simpler learned primitives.

G4 (waiting) was implemented as a rhythmic paw-lift by alternating the normal three-leg policy with rest phases. Across 15 lift–rest cycles, the policy maintained stable body height ($\sim$0.298 m $\pm$ 0.01 m) with $\sim$77\% lift success, though rest-phase reliability was lower due to manual settling. 

G5 (paw-offer/swing) was realized by combining the three-leg lift with an extended-reach variant. Under 31 rapid switching cycles, the extended policy achieved 100\% lift success with complete FL unloading ($\approx 0$ N) and stable height regulation ($\sim$0.306 m $\pm$ 0.003 m). 

Together, these results confirm that more socially expressive gestures can be composed from the trained primitives while preserving balance in simulation.

% Please explain this in simple words at the beginning of the paragraph.

\subsection{Ablation Study: The Role of Curriculum Learning} 

\begin{figure}[!h]
    \centering
    \captionsetup{font=small}
    \begin{subfigure}{0.95\columnwidth}
        \includegraphics[width=\linewidth]{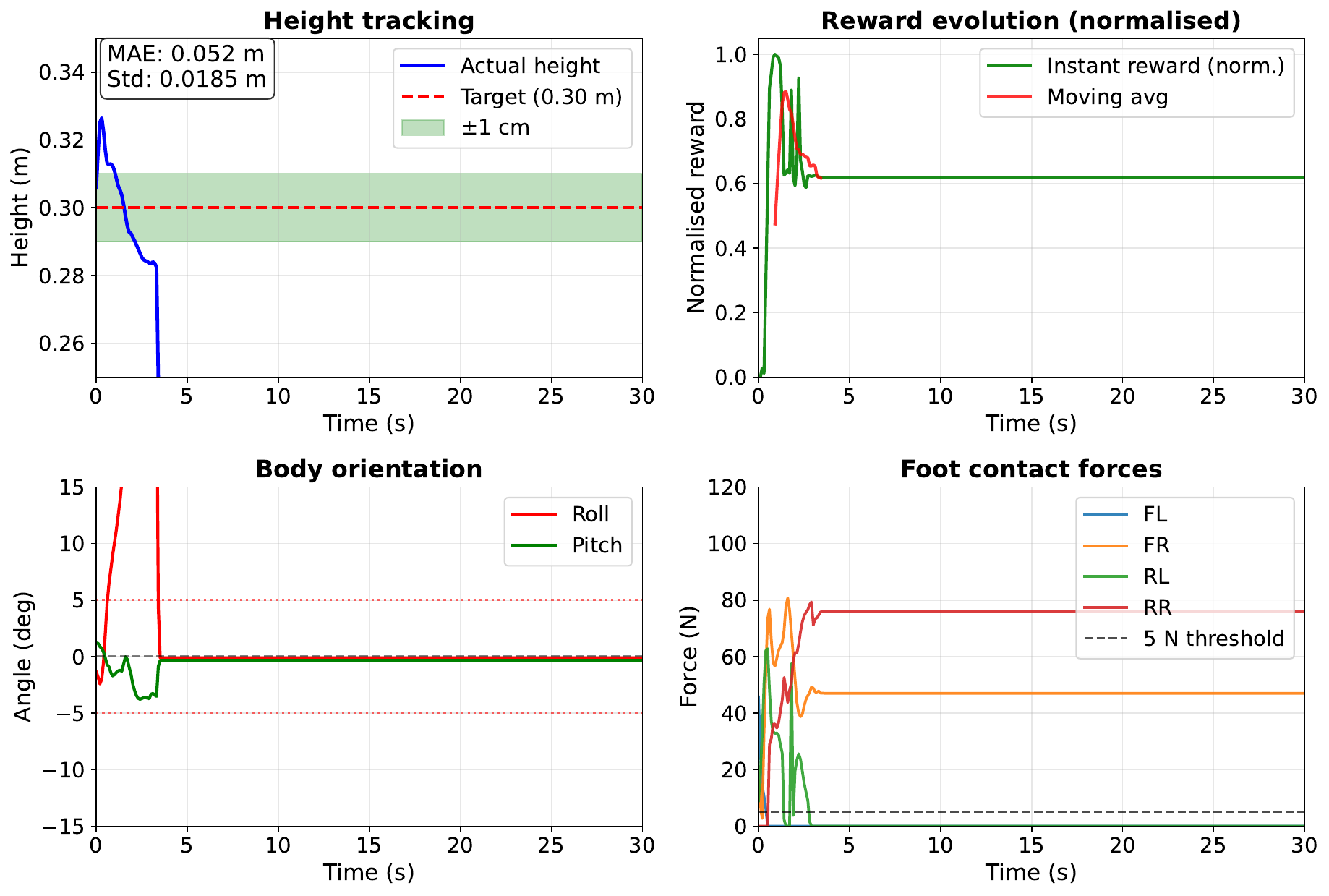}
        \caption{Without Curriculum Learning (Direct)}
        \label{fig:ablation_without_cl}
    \end{subfigure}
    \hfill
    \begin{subfigure}{0.95\columnwidth}
        \includegraphics[width=\linewidth]{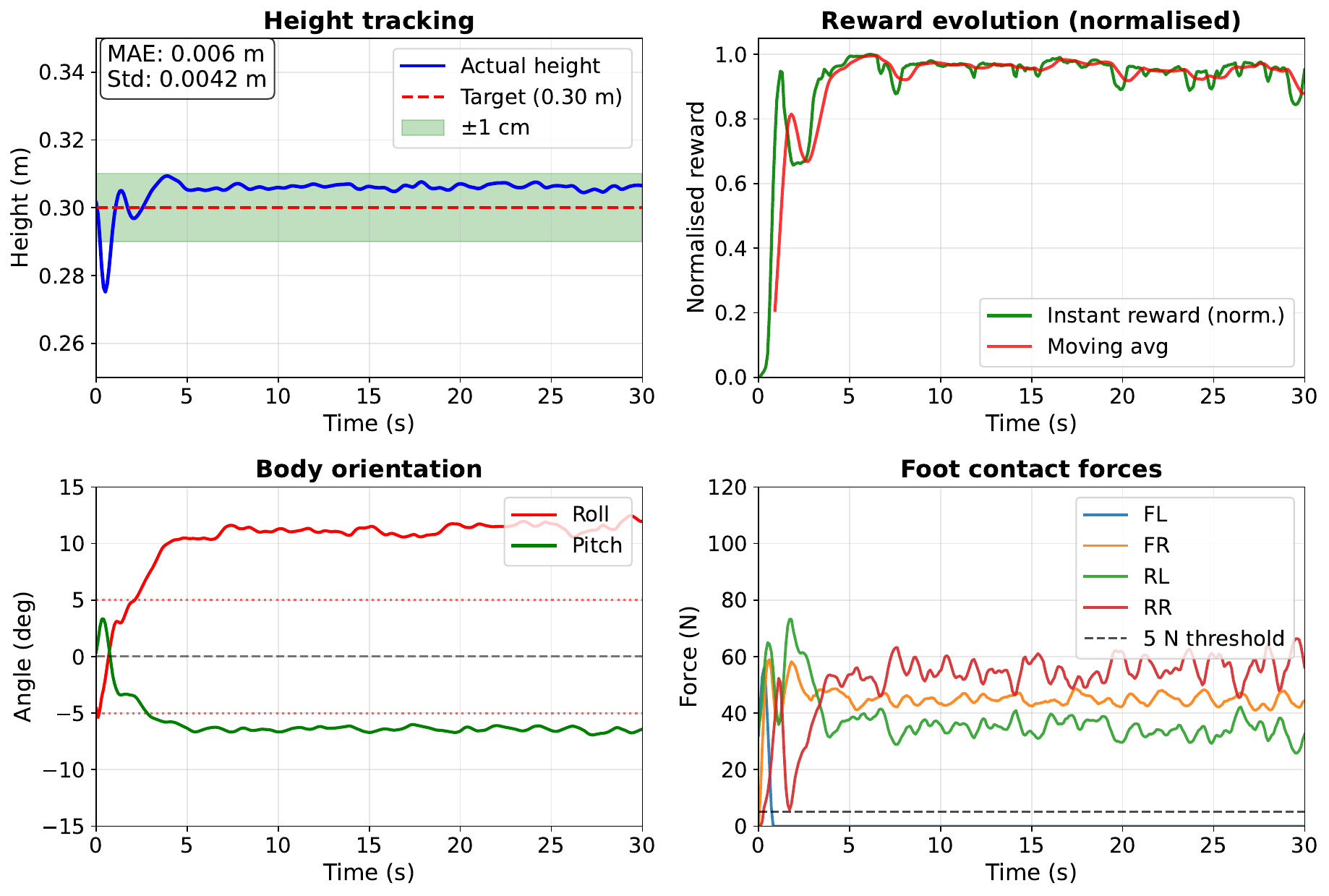}
        \caption{With Curriculum Learning (Our Method)}
        \label{fig:ablation_with_cl}
    \end{subfigure}
    \caption{Time-series data from the ablation study. (a) Direct training results in unstable behaviour, with erratic contact forces and orientation, leading to an early fall. (b) Our curriculum-based method achieves stable height tracking, consistent tripod contact forces (FL near zero), and a learned, steady roll angle for balance. 
For clarity of comparison, all reward curves are normalised to the range [0,\,1].}

    \label{fig:ablation_plots}
\end{figure}

The ablation study is conducted on the paw-lift gesture (FL leg raise). 
The results are shown in Fig.~\ref{fig:ablation_plots}, with (a) representing direct training without curriculum learning 
and (b) our proposed curriculum-based method.

Quantitative metrics (episode duration, contacts, height MAE, roll RMS) are summarised in Table~\ref{tab:ablation_results}.

For the policy trained with curriculum learning (Fig.~\ref{fig:ablation_plots}b), 
\emph{Height Tracking Performance} (top-left) shows rapid convergence to the target body height (\SI{0.30}{m}) with very small error (MAE $\approx$ \SI{0.006}{m}), indicating precise and stable control.
\emph{Reward Evolution} (top-right) stabilises quickly with a moving-average mean of $\approx 0.26$, confirming consistent learning progress.
\emph{Body Orientation Stability} (bottom-left) shows pitch near zero and a modest steady lean in roll (RMS $\approx 10.7^{\circ}$), reflecting a learned counter-balancing strategy rather than instability.
\emph{Foot Contact Forces} (bottom-right) indicate even sharing across the three support legs (FR, RL, RR) while the lifted FL leg remains near zero, demonstrating a successful three-leg stance with the intended paw-lift gesture.

In contrast, the policy trained without curriculum learning (Fig.~\ref{fig:ablation_plots}a) 
exhibits unstable behaviour. 
\emph{Height Tracking Performance} shows a rapid departure from the target with larger error (MAE $\approx$ \SI{0.013}{m}) and early termination.
\emph{Reward Evolution} oscillates heavily and does not stabilise within the short episode before failure.
\emph{Body Orientation Stability} shows roll drifting beyond the $\pm 5^{\circ}$ bound with no recovery, while pitch also deviates.
\emph{Foot Contact Forces} are erratic, with inconsistent load distribution and failure to unload the FL leg, leading to a fall after only about \SI{3}{s}.

\begin{table}[h!]
\centering
\captionsetup{font=small}
\caption{Ablation study results comparing training with and without Curriculum Learning (CL) for the three-leg paw-lift gesture.}
\label{tab:ablation_results}
\renewcommand{\arraystretch}{1.1}
\begin{tabularx}{\columnwidth}{@{}l X X@{}}
\toprule
\textbf{Metric} & \textbf{Without CL (Direct)} & \textbf{With CL (Our Method)} \\
\midrule
Episode Duration & \SI{3.3}{\second} (Early Term.) & \SI{30.0}{\second} (Full) \\
Avg. Ground Contacts & 2.6 / 4 & 3.0 / 4 \\
Height MAE & \SI{0.013}{\meter} & \SI{0.006}{\meter} \\
Roll RMS Stability & N/A (Unstable) & \SI{10.7}{\degree} (Stable Lean) \\
\bottomrule
\end{tabularx}
\end{table}

Overall, these results confirm that curriculum learning is essential for training complex social gestures: by gradually shaping the task, the agent first masters stability before learning the nuanced three-leg motion, whereas direct training fails to discover a viable strategy.

\subsection{Performance on Real Robot}
We focused real-robot evaluation on the front-left paw-lift (G1), other gestures were validated in simulation. We deployed the trained policies on the Unitree Go1 robot under low-level position control. The bias-free policy reproduced the intended front-left paw-lift gesture with clearer visual expression compared to the small-bias variant.

\begin{figure}[h]
\centering
\captionsetup{font=small}
\includegraphics[width=0.6\columnwidth]{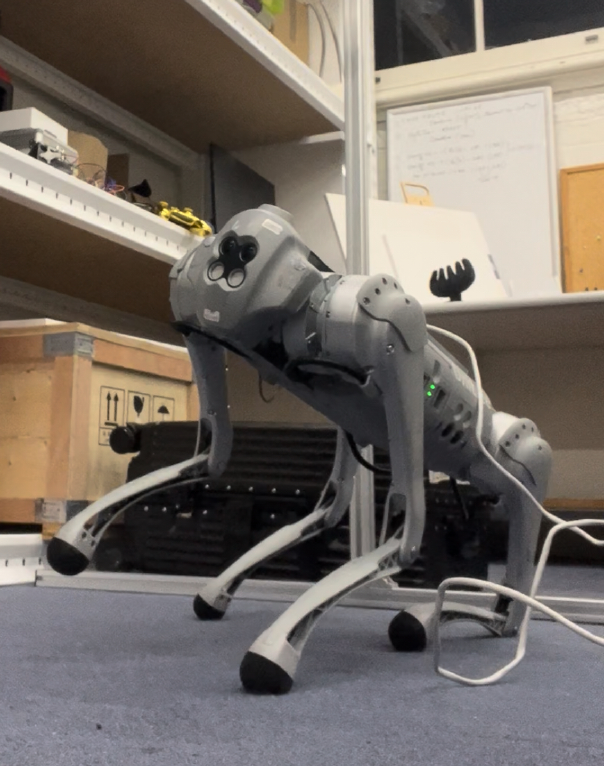}
\caption{Deployment of the FL-lift gesture on the Unitree Go1 robot in the real world.}
\label{fig:real}
\end{figure}

As shown in Fig.~\ref{fig:real}, the movement started around \SI{3.0}{\second}; the FL thigh joint gradually increased from +0.722 to \(\sim\)+0.731~rad, while the FL calf joint extended from \(-\)1.444 to \(\sim\)\(-\)1.634~rad, producing a visible and stable three-leg balance posture. The inference latency averaged \SI{26.97}{\milli\second} (min \SI{0.66}{\milli\second}, max \SI{84.3}{\milli\second}), confirming real-time feasibility within the \SI{20}{\hertz} control loop.

The bias-free policy yielded a cleaner and more symmetrical tripod stance, making the raised leg more visually distinct and socially expressive for human observers. In contrast, the small-bias version redistributed the load to FR/RL/RR and only slightly lifted FL, which appeared less natural.

A key hardware observation was that the RR leg showed knee flexion and occasional shank contact with the ground. This suggests that the discrepancy is not explained by a simple load distribution alone. It likely results from (i) joint compliance and mechanical backlash combined with conservative impedance gains, (ii) small pose offsets in the nominal posture that reduce RR ground clearance, and (iii) torque-balancing effects where modest vertical forces can create significant joint torques due to moment arms or transient peaks. Importantly, this behaviour may also reflect an internal balance mechanism embedded in the low-level Go1 controller, which adjusts the compliance and torque distribution in ways not captured by the simulator. This difference highlights a critical sim-to-real gap: the real robot may implicitly stabilise itself through hardware-level balance strategies absent in Isaac Gym.
These results confirm that the learned gestures are feasible on hardware and visually expressive, but also reveal that the lack of such balance control in simulation limits transferability. Introducing a simplified version of this mechanism into the simulator or compensating for it during training may be necessary to bridge this gap.

Demonstration videos of the robot’s performance in simulation and real-robot deployment are available online.
\footnote{Demonstration videos available at \url{https://tinyurl.com/reactive-go1}}

\section{Conclusion and Future Work} 

%Create a link with videos of the robto performance.
%Footnotes

This paper presented a senior-robot interaction framework for quadruped robots that combines user-friendly gesture-based control with a series of socially expressive robot behaviours. On the input side, we developed a perception interface using MediaPipe for hand gesture and head movement recognition, enabling older adults to control the robot naturally without handheld devices. On the output side, we introduced a set of socially expressive gestures, trained through a curriculum reinforcement learning framework, that enable the robot to communicate with the user. Simulation results in Isaac Gym confirmed the stability, robustness, and repeatability of these gestures, while deployment on the Unitree Go1 validated their feasibility in real hardware. Together, these contributions advance the role of quadruped robots as accessible and socially engaging companions for senior care.

In real-world deployment, the bias-free policy successfully reproduced the intended front-left leg lift with clear visual expression, but the observed amplitude was smaller than in simulation and a rear-right (RR) knee flexion with shank contact was present. These effects highlight key sim-to-real discrepancies, including hardware compliance and backlash, limited torque capacity under position control, small posture offsets that reduce ground clearance, and centre-of-pressure shifts that alter load distribution. Furthermore, the RR leg flexion may reflect an implicit balance control mechanism present in the Go1 hardware but absent in simulation, which could undermine the transferability of policies trained purely in Isaac Gym.

Looking forward, future work will focus on strengthening sim-to-real adaptation through several directions. First, we will explore residual or torque-based control together with hybrid impedance strategies to better capture the physical dynamics of the hardware. In addition, a simplified balance mechanism will be reimplemented in the simulator to approximate the Go1’s internal stabilization, thereby reducing the mismatch between simulated and real behaviours. Domain randomization will also be broadened to incorporate variations in compliance, friction, and latency, while on-device parameter identification with latency-aware observations will be employed to improve model fidelity. Beyond technical robustness, we aim to expand the socially expressive behaviours beyond front-leg gestures, potentially integrating multimodal cues such as sound or LED feedback, while maintaining slow and predictable dynamics suitable for senior users. Finally, systematic user studies with older adults will be conducted to evaluate gesture readability, comfort and accessibility, ensuring that the proposed framework translates into meaningful social interaction in real-world care contexts using a human-centred design approach.

%\section*{Acknowledgments}
%Make it anonymous at the time of submission
%Part of this research has been supported by the Scientia Development package PS46183-A (Dr Sandoval) and the internal grant to purchase the robot dog. 

%MAKE REFERENCES IN A NEW PAGE
%\newpage
%% This section was initially prepared using BibTeX.  The .bbl file was
%% placed here later
%\bibliography{publications}
%\bibliographystyle{named}
\bibliographystyle{named}
\balance
\bibliography{references}

\end{document}